\documentclass{article}

\usepackage[preprint, nonatbib]{neurips_2021}

\usepackage[utf8]{inputenc} % allow utf-8 input
\usepackage[T1]{fontenc}    % use 8-bit T1 fonts
\usepackage{hyperref}       % hyperlinks
\usepackage{url}            % simple URL typesetting
\usepackage{booktabs}       % professional-quality tables
\usepackage{amsfonts}       % blackboard math symbols
\usepackage{nicefrac}       % compact symbols for 1/2, etc.
\usepackage{microtype}      % microtypography
\usepackage{xcolor}         % colors

\usepackage{booktabs}
\usepackage{algorithm}
\usepackage{algpseudocode}
\usepackage{colortbl}
\usepackage{multirow}

\usepackage{amsmath}
\usepackage{amssymb}
\usepackage{graphicx,tikz}
\usetikzlibrary{patterns}
\usetikzlibrary{matrix,positioning}
\usepackage{pgfplots}
\usepgfplotslibrary{groupplots}
\pgfplotsset{compat=newest}
\DeclareMathOperator*{\argmax}{arg\,max}
\DeclareMathOperator*{\argmin}{arg\,min}

\usepackage{epstopdf}
\usepackage{epsfig}

\usepackage{float}

\title{Domain Adaptation without Model Transferring}

\author{%
  Kunhong~Wu, Yucheng~Shi, Yahong~Han \\
  College of Intelligence and Computing \\
  Tianjin University, Tianjin, China \\
  \texttt{ \{wkh16121, yucheng, yahong\}@tju.edu.cn }  \\
  \AND
  Yunfeng~Shao, Bingshuai~Li, Qi Tian \\
  Huawei Noah's Ark Lab\\
  Huawei Technologies\\
  \texttt{ \{shaoyunfeng, libingshuai, tian.qi1\}@huawei.com }  \\ }

\begin{document}

\maketitle

\begin{abstract}
  In recent years, researchers have been paying increasing attention to the threats brought by deep learning models to data security and privacy, especially in the field of domain adaptation. Existing unsupervised domain adaptation (UDA) methods can achieve promising performance without transferring data from source domain to target domain. However, UDA with representation alignment or self-supervised pseudo-labeling relies on the transferred source models. In many data-critical scenarios, methods based on model transferring may suffer from membership inference attacks and expose private data. In this paper, we aim to overcome a challenging new setting where the source models cannot be transferred to the target domain. We propose Domain Adaptation without Source Model, which refines information from source model. In order to gain more informative results, we further propose Distributionally Adversarial Training (DAT) to align the distribution of source data with that of target data. Experimental results on benchmarks of Digit-Five, Office-Caltech, Office-31, Office-Home, and DomainNet demonstrate the feasibility of our method without model transferring.
\end{abstract}

\section{Introduction}

Unsupervised domain adaptation (UDA) \cite{ganin2016domain, long2015learning, kim2019self, liang2019exploring} that uses the labeled source data to improve the accuracy on unlabeled target data becomes an important task because of its characteristic that reduces the workload of labeling. Meanwhile, the practical applications of existing deep learning methods in scenarios where data are decentralized and private, requires a level of trust association with data utilization and model training. To this end, there is a rising tendency of work on source-free unsupervised domain adaptation (source-free UDA) \cite{McMahan2017CommunicationEfficientLO,geyer2017differentially}, which attempts to achieve knowledge transfer from source domain to target domain without access to the source data.

There are two categories in existing source-free UDA methods: representation alignment and self-supervised pseudo-labeling methods. Source-free UDA methods based on representation alignment \cite{kd3a,li2020model} minimize the discrepancy between the source and target domains by aligning the representations or features of the data. Methods based on pseudo-labeling \cite{liang2020shot,kim2020domain} utilize models pre-trained on source domains to label the unlabeled data on the target domain. However, these two categories of source-free UDA methods both require model transferring between the source domains and the target domain. In other words, the model trained on the source domain needs to be handed over to the target domain for further operation \cite{liang2020shot,kd3a}, or models trained on the target domain are required to deliver to the source domain for evaluation.

Transferring of models provides higher security than transferring of source data, gradients or features \cite{mcmahan2017communication}. However, once the source model is intercepted during transferring, its training data and model information may suffer from membership inference attack \cite{shokri2017membership} or model extraction attack \cite{tramer2016stealing}, and thus lead to privacy leakage, even if only model is transferred \cite{truex2019demystifying}. The solution to block the risk of privacy leakage in practical source-free UDA applications is to completely abandon model transferring. However, in order to simulate and approximate the feature space of the source model, we require another carrier for supervision information between source and target domains.

We propose a domain adaptation framework that copes with the challenging setting of source-free UDA where the model cannot be transferred between source and target domains. Moreover, to avoid the impact of membership inference attack or model distillation based on confidence information, we further limit the output of source model to hard labels (labels without logits or confidence). Instead of using source data, target data or source model, SFUDA uses another dataset as the carrier of supervision information between domains. SFUDA first uses the source models pre-trained on multiple source domains, and then uses the returned hard labels with the other data to initialize the target model.

However, due to the large gap in data distribution, category, and image shape between another dataset and the source data, it is difficult to obtain unbiased supervision information from the source model by only with another dataset. Aiming at obtaining more direct and unbiased supervision information from source domain, we need to adjust the distribution the other data to approximate that of the target data. We use adversarial training \cite{Tramr2017EnsembleAT}, a strategy that adds noises on the training data to improve the generalization ability of model in different data distributions, to align another data with target data. Existing adversarial training methods use object function based on labels, which are noisy and biased on the target model. Instead, we propose Distributionally Adversarial Training (DAT) to reduce the KL divergence of feature distributions between other data and the target data on the target model. We  the source models with adversarial examples generated by DAT, and retrain the target model with adversarial examples and corresponding queried labels. In addition, we use clustering algorithm \cite{liang2020shot} to label target data and refine all the pseudo-labels with DEPICT algorithm \cite{ghasedi2017deep}.
%In summary, the target model in SFUDA is initialized on third-party data, retrained on adversarial examples generated by DAT, and fine-tuned by pseudo-labeled target data.

During the entire process of SFUDA, neither source data or model is transferred to the target domain, nor any target domain data is used to  the source model. The feature and gradients of the source and target data are strictly limited to their respective domains during the  process, which fully guarantees the data privacy. In addition, the setting that the source model in SFUDA only outputs hard labels is consistent with many systems that encapsulate machine learning models as APIs or cloud services, thus broadening the practicability of SFUDA in privacy related fields. The experimental results on unsupervised domain adaptation datasets reveal that the proposed SFUDA achieves comparable performance without transferring of source and target models. We further use membership inference attack \cite{shokri2017membership} to verify the data privacy of SFUDA without model transferring, compared to other source-free UDA methods.

\section{Related Work}

\subsection{Various Domain Adaptation Settings}
We compare four domain adaptation settings. In the supervised domain adaptation hypothesis, both labeled source and target data can be accessed \cite{tzeng2017adversarial}. Compared with supervised domain adaptation, UDA \cite{liang2019exploring, liang2018aggregating} does not use the label of target domain and therefore reduces the labeling workload. In order to improve the data privacy on different organizations or devices, the source-free UDA prohibits the transferring of data between domains.

\subsection{Source-free Unsupervised Domain Adaptation}
There are two categories of methods in Source-free UDA: representation alignment methods and self-supervised pseudo-labeling methods. As a representation alignment-based method, FADA enhances knowledge transfer through dynamic attention mechanism and feature disentanglement. Specifically, FADA trains a feature extractor and classifier on each source domain and aggregates their gradients in the target model. The weight of each source model is assigned according to the improvement of the feature extractor on the target domain. Other representation alignment methods such as KD3A \cite{kd3a} and Model Adaptation \cite{li2020model} use knowledge transfer and style transfer to provide supervision information for the target model. SHOT \cite{liang2020shot} uses pseudo-labeling to achieve domain adaptation under the condition that only source model can be obtained. Specifically, SHOT trains feature extractor and classifier of the source domain and send them to the target domain for pseudo-labeling. SHOT further train the target model by screening out target data with the highest pseudo-label confidence. Other pseudo-labeling methods, such as SFDA \cite{kim2020domain}, adopt distance-based confidence and remove unreliable target data to improve the quality of pseudo-labels.

However, source-free UDA methods based on model transferring may suffer from model extraction attack \cite{tramer2016stealing} or membership inference attack \cite{shokri2017membership} during the model transferring process. The model extraction attacks try to obtain the internal information of the model by continuously ing the machine learning model. The membership inference attacks establish connection between the input and output of the machine learning model, and infer the training data according to certain search strategy. It will pose serious threat to the data privacy of each domain if these two methods are utilized to attack the source-free UDA methods based on model transferring.

The IterNLL \cite{zhang2021unsupervised}, UB2DA \cite{deng2021universal} and Dis-tune \cite{liang2021distill} methods treat the source model as an input-output interface, and use noisy label learning and knowledge distillation methods to improve the pseudo-labeling quality of the source model for unlabeled target data. There are three main differences on settings between SFUDA and these three methods. First of all, setting of SFUDA does not allow any model transferring. Secondly, the source model can only output hard labels in the  process under our setting, while the cores of these three methods are based on the soft label output by the source model. Most importantly, IterNLL, UB2DA and Dis-tune directly use the target data to , which is not allowed under our setting considering the data privacy. Thus the setting of proposed SFUDA is more privacy-preserving while more challenging than the existing source-free UDA settings.

\section{Domain Adaptation without Model Transferring}
%In this section, we firstly describe the notation and black-box source-free UDA setting. Then we detail the  process based on third-party dataset and Distributionally Adversarial Training of SFUDA. Finally, we introduce two other probe strategies for comparison in experiments.
%
%\subsection{Notation}
SFUDA considers a source-free UDA setting. Given the data and labels of $N$ source domains $\mathbb{D}_{S}=\{ \mathbb{D}_{S_i} \}_{i=1}^N, \ \mathbb{D}_{S_i} = \{ (x_{S_i}^j, y_{S_i}^j) \}_{j=1}^{n_{S_i}}$ with $n_{S_i}$ samples for each source domain and the unlabeled data on a target domain $\mathbb{D}_T = \{ (x_T^j) \}_{j=1}^{n_T}$ with $n_T$ samples, the goal is to maximize the accuracy of a target model on the test set of the target domain:
\begin{equation}\label{goal}
  \max_{M_T}\mathop{E}\limits_{\substack{(x,y)\sim \mathbb{D}_T}}[M_T(x)=y],
\end{equation}
where $M_T: X^{W\times H\times C}\rightarrow Y^K$ is the target model, $X$ represents the input space with dimension of $Width\times Height\times Channel$, $Y$ represents the classification space with $K$ categories. All models in this paper are composed of two parts: feature extractor $g$ and classifier $h$:
\begin{equation}\label{g_h}
  M(x) = \argmax \sigma(h(g(x))),
\end{equation}
where $\sigma$ denotes softmax function. The feature extractor $g$ receives images as input and outputs the feature map. The output of $g$ is passed to the classifier $h$, which outputs logits of the same dimension as the number of categories. Finally, the probability vector is obtained through the softmax. The index of maximum value is taken as the predict category. Source model $M_{S_i}$ trained on each of $N$ source domains cannot be transferred to the target domain. The  between the source domain and the target domain is allowed, but target data cannot be input to any of $M_{S_i}$, nor can the target model be evaluated on the $\mathbb{D}_S$. The source data also cannot be directly sent to the target model for :
\begin{equation}
  y_{} = M(x_{}), \quad s.t. \quad x_{} \notin [\mathbb{D}_S, \mathbb{D}_T], \ M \in [M_{S_1}, M_{S_2}, \cdots, M_{S_N}, M_T].
\end{equation}

\subsection{Distributionally Adversarial Training}

The domain gap between $\mathbb{D}_E$ and $\mathbb{D}_T$ is usually much larger than that between $\mathbb{D}_S$ and $\mathbb{D}_T$. In addition, due to the labeling bias of $M_S$ to $\mathbb{D}_E$, the performance of $M_T$ initialized with $\mathbb{D}_E$ may be sub-optimal. IterNLL, UB2DA and Dis-tune adjust the sampling and distilling strategies according to the confidence of the target data on the transferred source model to reduce the impact of noisy labels on the target model, which are not available under our source-free UDA setting. We instead use target data and adversarial examples based on another dataset to fine-tune the target model. We first use the target model to pseudo-label the target data, and adopt DEPICT structure again for label refinement:
\begin{equation}\label{target_depict}
  p_{jk}^T = \sigma_k (h_T(g_T(x_T^j))), \quad  \hat{q}^T_{jk} = \frac{p_{jk}^T \/ (\sum_{j^\prime}p^T_{j^\prime k})^{\frac{1}{2}}}{\sum_{k^\prime}p_{jk^\prime}^T \/ (\sum_{j^\prime}p_{j^\prime k^\prime}^T)^{\frac{1}{2}}},
\end{equation}
where $\hat{q}^T_{jk}$ represents the refined confidence of the $k_{th}$ category on the $j_{th}$ target data of the target model. $h_T$ and $g_T$ denote the classifier and feature extractor of target model, respectively. Inspired by SHOT \cite{liang2020shot}, we cluster target data with features and refined pseudo-labels and readjust the pseudo-labels according to the distance between the features of the target data and the clustering centroid:
\begin{equation}\label{shot_adjust}
  \rho_k = \frac{\sum_j ( \hat{q}^T_{jk} \cdot g_T(x_T^j))}{\sum_j \hat{q}^T_{jk}}, \quad \hat{y}_T = \argmin_k Cosine(g_T(x_T), \rho_k),
\end{equation}
where $\rho_k$ denotes the cluster centroid of the $k_{th}$ category on the target data using the feature extractor of the target model, $\hat{y}_T$ represents the readjusted pseudo-label by calculating the cosine distance between each target data and cluster centroid. We fine-tune the target model $M_T$ using target data $x_T$ and its pseudo-label based on label refinement and clustering:
\begin{equation}\label{target_loss}
  L_{target} = \frac{1}{n_T} \sum^{n_T}_{j=1} CrossEntropy(M_T(x_T^j), \hat{y}_T).
\end{equation}
\begin{algorithm}[t]
\caption{Domain Adaptation without Model Transferring} %算法的名字
\hspace*{0.02in} {\bf Input:} %算法的输入， \hspace*{0.02in}用来控制位置，同时利用 \\ 进行换行
$N$ source domains $\mathbb{D}_{S}=\{ \mathbb{D}_{S_i} \}_{i=1}^N$ and target domain $\mathbb{D}_T = \{ (x_T^j) \}_{j=1}^{n_T}$  \\
\hspace*{0.42in} Another dataset $\mathbb{D}_E$, initial parameters $\theta_T$ for target model $M_T$ and optimizer $Opt$\\
\hspace*{0.02in} {\bf Output:} %算法的结果输出
Target model $M_T$
\begin{algorithmic}[1]
\State Train source models $\{ M_{S_i} \}_{i=1}^N$  on each source domain  $\mathbb{D}_{S_i} = \{ (x_{S_i}^j, y_{S_i}^j) \}_{j=1}^{n_{S_i}}$;
\State Obtain averaged confidence on another data from source models $\sigma(\frac{1}{N} \sum_{i=1}^{N}  h_{S_i}(g_{S_i}(x_E)))$;
\State Perform label refinement on confidence according to Eqn. \eqref{depict};
\State Return the hard pseudo-label to target model $\hat{M}_S(x_E)$;
\State Initialize target model with other data using Eqn. \eqref{init_loss}: $\theta_T \leftarrow Opt(L_{other}, \theta_T)$;
\State Input target data into target model and perform label refinement according to Eqn. \eqref{target_depict};
\State Cluster features of target data according to Eqn. \eqref{shot_adjust} and generate pseudo-labels on target data;
\State Fine-tune the target model with target data using Eqn. \eqref{target_loss}: $\theta_T \leftarrow Opt(L_{target}, \theta_T)$;
\State Iteratively generate adversarial examples $x^\prime_E$ based on KL divergence between the other data and target data on target feature extractor according to Eqn. \eqref{DAT};
\State Average confidence on adversarial examples from source models $\sigma(\frac{1}{N} \sum_{i=1}^{N}  h_{S_i}(g_{S_i}(x^\prime_E)))$;
\State Return the hard pseudo-label of $x^\prime_E$ after label refinement in Eqn. \eqref{depict} to target model $\hat{M}_S({x_E^\prime})$;
\State Retrain target model using adversarial examples using Eqn. \eqref{DAT_loss}: $\theta_T \leftarrow Opt(L_{DAT}, \theta_T)$;
\State Fine-tune the target model $M_T$ again according to Eqn. \eqref{shot_adjust} and \eqref{target_loss};
\end{algorithmic}
\label{alg:1}
\end{algorithm}

So far, the only source of supervision information obtained is the hard label of the another dataset on the source model. To obtain more direct and unbiased supervision information from the source domain during the  process, we propose Distributionally Adversarial Training on another dataset to reduce the distribution difference between $x_E$ and $x_T$. We firstly construct adversarial examples based on KL divergence of target feature extractor $g_T$ between $x_E$ and $x_T$:
\begin{equation}\label{DAT}
  x^\prime_{E(0)} = x_E, \quad x^\prime_{E(t+1)} = Clip_x \{ x^\prime_{E(t)} - \mu\cdot \bigtriangledown_{x^\prime_{E(t)}} D_{kl}[g_T(x_T), g_T(x^\prime_{E(t)})] \},
\end{equation}
where $x^\prime_{E(t)}$ denotes the adversarial example after $t$ iterations, $Clip_x$ limit the adversarial example within licit range, $\mu$ is the step size of the iterative adversarial attack. During DAT, we randomly sample two batches of data from the target domain and another dataset, and minimize the KL divergence of their features by adding noise to the another data, thereby guiding $x_E$ closer to $x_T$. The target model and data are all frozen in DAT, and only the adversarial examples based on $x_E$ are optimizable. After generating adversarial examples $x^\prime_E$ on the entire another dataset, we once again  the source model to obtain hard pseudo-labels and retrain the target model:
\begin{equation}\label{DAT_loss}
  L_{DAT} = \frac{1}{n_E} \sum^{n_E}_{j=1} CrossEntropy(M_T({x^\prime_E}^j), \hat{M}_S({x^\prime_E}^j)).
\end{equation}
The label refinement for adversarial examples is consistent with Eqn. \eqref{depict}. Finally, we utilize pseudo-label refinement of the target data (Eqn. \eqref{target_depict}) and clustering (Eqn. \eqref{shot_adjust} and \eqref{target_loss}) to fine-tune the retrained target model as the final model. Algorithm 1 details the whole process of SFUDA.

\subsection{Other Strategies}
\label{sec_other_}
In addition to SFUDA, we introduce two strategies without model transferring, which are used for comparison in the experimental part.

\textbf{Centroid based  (CP).} When there is no label to rely on, the cluster centroid usually contains more features of this domain \cite{liang2020shot}. Therefore, we use $\textit{K}$-means to cluster the source domain data:
\begin{equation}\label{centroid}
  \min \limits_{\substack{\eta \\ x_T \in \mathbb{D}_T}} \sum_{i=1}^{K} \sum_{x_T^i \in \varphi_i} \| x_T^i - \eta_i \|_2^2, \quad \eta_i = \frac{1}{|\varphi_i|}\sum_{x_T^i \in \varphi_i}x_T^i, \quad x_{train}^{CP} = [\eta_1, \eta_2, \cdots, \eta_K],
\end{equation}
where $\eta_i$ is the mean vector of the i-th cluster $\varphi_i$, $K$ is the number of categories.

\textbf{Gaussian Noise based  (GNP).} We also try to  the source model using Gaussian noises:
\begin{equation}\label{Gauss_}
  x_{train}^{GNP} \sim \mathcal{N}_{(W,H,C)}(0,1),
\end{equation}
where $\mathcal{N}_{(W,H,C)}$ denotes the multivariate normal distribution with dimension of $W\times H\times C$.

% Table generated by Excel2LaTeX from sheet 'digit5'
\begin{table}[t]
  \centering
  \caption{Accuracies (\%) on Digit-Five for multi-source UDA}
    \small
    \begin{tabular}{c|c|cccccc}
    \toprule
    \multicolumn{1}{c|}{Standards} & Methods & mnist & mnistm & svhn  & syn   & usps  & \cellcolor[rgb]{ .851,  .851,  .851}Avg \\
    \midrule
    \multicolumn{1}{c|}{\multirow{2}[0]{*}{W/o DA}} & Source Only & 97.5  & 67.0  & 63.0  & 76.5  & 93.1  & \cellcolor[rgb]{ .851,  .851,  .851}79.4  \\
    \multicolumn{1}{c|}{} & Oracle & 99.6  & 97.8  & 92.8  & 98.7  & 99.4  & \cellcolor[rgb]{ .851,  .851,  .851}97.7  \\
    \midrule
    \multicolumn{1}{c|}{Source-free} & SHOT  & 98.2  & 80.2  & 84.5  & 91.1  & 97.1  & \cellcolor[rgb]{ .851,  .851,  .851}90.2  \\
    \multicolumn{1}{c|}{UDA} & FADA  & 91.4  & 62.5  & 50.5  & 71.8  & 91.7  & \cellcolor[rgb]{ .851,  .851,  .851}73.6  \\
    \midrule
    Our \& & CP    & 53.1  & 83.2  & 56.1  & 89.2  & 37.7  & \cellcolor[rgb]{ .851,  .851,  .851}63.9  \\
    Source-free & GNP   & 9.9   & 9.0   & 10.0  & 15.8  & 15.9  & \cellcolor[rgb]{ .851,  .851,  .851}12.1  \\
    UDA   & SFUDA (ours) & \textbf{98.7 } & \textbf{77.1 } & \textbf{65.1 } & \textbf{90.4 } & \textbf{96.7 } & \cellcolor[rgb]{ .851,  .851,  .851}\textbf{85.6 } \\
    \bottomrule
    \end{tabular}%
  \label{table_digit}%
\end{table}%

% Table generated by Excel2LaTeX from sheet 'office-caltech'
\begin{table}[t]
  \centering
  \caption{Accuracies (\%) on Office-Caltech for multi-source UDA}
    \small
    \begin{tabular}{c|c|ccccc}
    \toprule
    \multicolumn{1}{c|}{Standards} & Methods & amazon & caltech & dslr  & webcam & \cellcolor[rgb]{ .851,  .851,  .851}Avg \\
    \midrule
    \multicolumn{1}{c|}{\multirow{2}[2]{*}{W/o DA}} & Source Only & 94.5  & 92.3  & 98.1  & 97.6  & \cellcolor[rgb]{ .851,  .851,  .851}95.6  \\
    \multicolumn{1}{c|}{} & Oracle & 93.8  & 94.7  & 100.0  & 100.0  & \cellcolor[rgb]{ .851,  .851,  .851}97.1  \\
    \midrule
    \multicolumn{1}{c|}{UDA} & DAN   & 91.6  & 89.2  & 99.1  & 99.5  & \cellcolor[rgb]{ .851,  .851,  .851}94.8  \\
    \midrule
    \multicolumn{1}{c|}{Source-free} & SHOT  & 95.4  & 94.6  & 98.4  & 98.9  & \cellcolor[rgb]{ .851,  .851,  .851}96.9  \\
    \multicolumn{1}{c|}{UDA} & FADA  & 84.2  & 88.7  & 87.1  & 88.1  & \cellcolor[rgb]{ .851,  .851,  .851}87.0  \\
    \midrule
    Our \& & CP    & 92.9  & 61.6  & 98.1  & 93.9  & \cellcolor[rgb]{ .851,  .851,  .851}86.6  \\
    Source-free & GNP   & 11.0  & 16.1  & 17.2  & 20.5  & \cellcolor[rgb]{ .851,  .851,  .851}16.2  \\
    UDA   & SFUDA (ours) & \textbf{95.1 } & \textbf{94.5 } & \textbf{100.0 } & \textbf{99.3 } & \cellcolor[rgb]{ .851,  .851,  .851}\textbf{97.2 } \\
    \bottomrule
    \end{tabular}%
  \label{table_office_caltech}%
\end{table}%

% Table generated by Excel2LaTeX from sheet 'office'
\begin{table}[t]
  \centering
  \caption{Accuracies (\%) on Office-31 for single-source UDA}
    \small
    \begin{tabular}{c|c|ccccccc}
    \toprule
    \multicolumn{1}{c|}{Standards} & Methods & A -> D & A -> W & D -> A & D -> W & W -> A & W -> D & \cellcolor[rgb]{ .851,  .851,  .851}Avg \\
    \midrule
    \multicolumn{1}{c|}{\multirow{2}[2]{*}{W/o DA}} & Source Only & 81.3  & 74.3  & 61.7  & 94.7  & 62.9  & 97.6  & \cellcolor[rgb]{ .851,  .851,  .851}78.8  \\
    \multicolumn{1}{c|}{} & Oracle & 100.0  & 100.0  & 87.2  & 100.0  & 87.2  & 100.0  & \cellcolor[rgb]{ .851,  .851,  .851}95.7  \\
    \midrule
    \multicolumn{1}{c|}{\multirow{2}[2]{*}{UDA}} & DANN  & 79.7  & 82.0  & 68.2  & 96.9  & 67.4  & 99.1  & \cellcolor[rgb]{ .851,  .851,  .851}82.2  \\
    \multicolumn{1}{c|}{} & DAN   & 78.6  & 80.5  & 63.6  & 97.1  & 62.8  & 99.6  & \cellcolor[rgb]{ .851,  .851,  .851}80.4  \\
    \midrule
    \multicolumn{1}{c|}{Source-free} & SHOT  & 94.4  & 88.7  & 75.1  & 98.4  & 73.2  & 99.6  & \cellcolor[rgb]{ .851,  .851,  .851}88.2  \\
    \multicolumn{1}{c|}{UDA} & Dis-tune & 91.0  & 85.1  & 72.4  & 98.1  & 73.1  & 98.7  & \cellcolor[rgb]{ .851,  .851,  .851}86.4  \\
    \midrule
    Our \& & CP    & 77.5  & 61.4  & 62.4  & 83.8  & 62.4  & 86.4  & \cellcolor[rgb]{ .851,  .851,  .851}72.3  \\
    Source-free & GNP   & 4.1   & 4.9   & 3.6   & 4.3   & 2.8   & 4.3   & \cellcolor[rgb]{ .851,  .851,  .851}4.0  \\
    UDA   & SFUDA (ours) & \textbf{89.4 } & \textbf{88.4 } & \textbf{61.4 } & \textbf{93.3 } & \textbf{60.7 } & \textbf{94.8 } & \cellcolor[rgb]{ .851,  .851,  .851}\textbf{81.3 } \\
    \bottomrule
    \end{tabular}%
  \label{table_office_31}%
\end{table}%

\section{Experiments}
\subsection{Experimental Setup}
\label{sec_setting}

SFUDA is tested on Digit-Five \cite{zhao2020multi}, Office-Caltech \cite{gong2012geodesic}, Office-31, Office-Home and DomainNet \cite{peng2019moment} datasets. Digit-Five dataset contains 5 domains: MNIST-M, MNIST, SYN, USPS and SVHN. The Office-Caltech dataset contains 4 domains: Caltech(C), Amazon(A), WebCam(W) and DSLR(D). The Office-31 dataset contains 3 domains: Amazon(A), WebCam(W) and DSLR(D). The Office-Home dataset contains 4 domains: Artistic images(A), Clip Art(C), Product images(P) and Real-World images(R). The most challenging DomainNet dataset contains 6 domains: Clipart(C), Infograph(I), Painting(P), Quickdraw(Q), Real(R) and Sketch(S).

We use the validation set of ImageNet \cite{russakovsky2015imagenet} with 1000 categories and 50000 images as the another dataset. ImageNet and the above datasets are different in the number of categories and data distribution. As for the structure of DNNs, we adopt the classic configuration: 3-layer CNN for Digit-Five \cite{kd3a}, ResNet-50 \cite{he2016deep} for the others. The learning rate of SGD optimizer is set to $lr = 1e-3$. The batch size for Digit-Five is 200, the others are 64. The iteration number for DAT is set to 5. The step size for iterative adversarial attack $\mu$ is set to 5.

We compare the SFUDA with seven different methods on different target domains. Among them, `Oracle' indicates the accuracy obtained by directly training on the target data $\mathbb{D}_{T}$. `Source Only' refers to the accuracy of the ensembled source models. `GNP' and `CP' represent  strategies introduces in Section \ref{sec_other_} based on Gaussian noise and centroid, respectively. We also report the performance of two UDA methods DANN \cite{ganin2015unsupervised} and DAN \cite{long2015learning}, as well as three source-free UDA methods, SHOT \cite{liang2020shot} and Dis-tune \cite{liang2021distill}. Note that these methods allow model transferring from source domain to the target domain. The `standard' column appears in tables indicates the setting to which the each method belongs. All experiments are conducted on 8 RTX 3090 GPU.

% Table generated by Excel2LaTeX from sheet 'office-home'
\begin{table}[t]
  \centering
  \caption{Accuracies (\%) on Office-Home for single-source UDA}
    \small
    \tabcolsep=0.05cm
    \begin{tabular}{c|ccccccccccccc}
    \toprule
    Methods & A -> C & A -> P & A -> R & C -> A & C -> P & C -> R & P -> A & P -> C & P -> R & R -> A & R -> C & R -> P & \cellcolor[rgb]{ .851,  .851,  .851}Avg \\
    \midrule
    Source Only & 44.9  & 65.7  & 73.7  & 52.8  & 60.6  & 65.0  & 52.7  & 40.9  & 73.5  & 64.9  & 45.5  & 76.9  & \cellcolor[rgb]{ .851,  .851,  .851}59.8  \\
    Oracle & 79.2  & 95.9  & 85.8  & 74.1  & 95.9  & 85.8  & 74.1  & 79.2  & 85.8  & 74.1  & 79.2  & 95.9  & \cellcolor[rgb]{ .851,  .851,  .851}83.7  \\
    \midrule
    DANN  & 45.6  & 59.3  & 70.1  & 47.0  & 58.5  & 60.9  & 46.1  & 43.7  & 68.5  & 63.2  & 51.8  & 76.8  & \cellcolor[rgb]{ .851,  .851,  .851}57.6  \\
    DAN   & 43.6  & 57.0  & 67.9  & 45.8  & 56.5  & 60.4  & 44.0  & 43.6  & 67.7  & 63.1  & 51.5  & 74.3  & \cellcolor[rgb]{ .851,  .851,  .851}56.3  \\
    \midrule
    SHOT  & 57.2  & 76.5  & 80.4  & 67.8  & 76.8  & 78.2  & 67.2  & 54.3  & 82.4  & 71.9  & 58.8  & 82.9  & \cellcolor[rgb]{ .851,  .851,  .851}71.2  \\
    Dis-tune & 52.9  & 78.3  & 81.5  & 65.3  & 76.1  & 77.8  & 62.4  & 50.3  & 81.8  & 70.5  & 55.9  & 84.1  & \cellcolor[rgb]{ .851,  .851,  .851}69.7  \\
    \midrule
    CP    & 28.4  & 41.6  & 45.3  & 33.2  & 36.3  & 45.1  & 30.4  & 29.8  & 42.4  & 44.2  & 31.3  & 48.9  & \cellcolor[rgb]{ .851,  .851,  .851}38.1  \\
    GNP   & 2.7   & 2.1   & 2.2   & 2.7   & 2.5   & 2.7   & 2.0   & 2.9   & 2.5   & 2.6   & 2.4   & 2.2   & \cellcolor[rgb]{ .851,  .851,  .851}2.5  \\
    SFUDA (ours) & \textbf{50.2 } & \textbf{73.6 } & \textbf{77.5 } & \textbf{60.1 } & \textbf{70.5 } & \textbf{73.1 } & \textbf{62.7 } & \textbf{49.8 } & \textbf{78.6 } & \textbf{67.4 } & \textbf{51.9 } & \textbf{79.9 } & \cellcolor[rgb]{ .851,  .851,  .851}\textbf{66.3 } \\
    \bottomrule
    \end{tabular}%
  \label{table_office_home}%
\end{table}%

% Table generated by Excel2LaTeX from sheet 'domain-net'
\begin{table}[H]
  \centering
  \caption{Accuracies (\%) on DomainNet for multi-source UDA}
    \small
    \begin{tabular}{c|cccccccc}
    \toprule
    \multicolumn{1}{c|}{Standards} & Methods & clipart & infograph & painting & quickdraw & real  & sketch & \cellcolor[rgb]{ .851,  .851,  .851}Avg \\
    \midrule
    \multicolumn{1}{c|}{\multirow{2}[1]{*}{W/o DA}} & Source Only & 62.0  & 22.4  & 51.8  & 10.2  & 67.1  & 50.2  & \cellcolor[rgb]{ .851,  .851,  .851}44.0  \\
    \multicolumn{1}{c|}{} & Oracle & 76.2  & 38.9  & 71.0  & 69.7  & 82.7  & 68.5  & \cellcolor[rgb]{ .851,  .851,  .851}67.8  \\
    \midrule
    \multicolumn{1}{c|}{Source-free} & SHOT  & 61.7  & 22.2  & 52.6  & 12.2  & 67.7  & 48.6  & \cellcolor[rgb]{ .851,  .851,  .851}44.2  \\
    \multicolumn{1}{c|}{UDA} & FADA  & 45.3  & 16.3  & 38.9  & 7.9   & 46.7  & 26.8  & \cellcolor[rgb]{ .851,  .851,  .851}30.3  \\
    \midrule
    Our \& & CP    & 0.6   & 2.4   & 15.0  & 1.0   & 8.2   & 1.1   & \cellcolor[rgb]{ .851,  .851,  .851}4.7  \\
    Source-free & GNP   & 0.2   & 0.2   & 0.1   & 0.1   & 0.3   & 0.3   & \cellcolor[rgb]{ .851,  .851,  .851}0.2  \\
    UDA   & SFUDA (ours) & \textbf{45.3 } & \textbf{17.6 } & \textbf{45.2 } & \textbf{5.6 } & \textbf{63.9 } & \textbf{36.7 } & \cellcolor[rgb]{ .851,  .851,  .851}\textbf{35.7 } \\
    \bottomrule
    \end{tabular}%
  \label{table_domainnet}%
\end{table}%

% Table generated by Excel2LaTeX from sheet 'Abalation'
\begin{table}[t]
  \centering
  \caption{Ablation study of three parts of supervision information in SFUDA. The accuracy on each dataset is averaged on all the multi-domain adaptation tasks.}
    \small
    \tabcolsep=0.07cm
    \begin{tabular}{c|ccccc}
    \toprule
    Methods & DomainNet & Office-31 & Office-Caltech & Office-Home & Digit-Five \\
    \midrule
    Source Only & \textbf{44.0}  & 85.5  & 95.6  & 68.9  & 79.4 \\
    $L_{another}$ & 30.8  & 84.5  & 96.3  & 67.4  & 87.1 \\
    $L_{another} + L_{DAT}$ & 35.4  & 86.6  & 96.9  & 69.9  & 93.9 \\
    $L_{another} + L_{DAT} + L_{target}$ (SFUDA)   & 35.7 & \textbf{87.2} & \textbf{97.2} & \textbf{70.2} & \textbf{96.7} \\
    \bottomrule
    \end{tabular}%
  \label{table_ablation_study}%
\end{table}%

% Table generated by Excel2LaTeX from sheet 'membership'
\begin{table}[t]
  \centering
  \caption{Comparison of the membership inference attack model's judgement accuracy between source model on DomainNet and initialized model of SFUDA using ImageNet dataset.}
    \small
    \begin{tabular}{c|c|cccccc}
    \toprule
          & Model & clipart & infograph & painting & quickdraw & real  & sketch \\
    \midrule
    \multirow{2}[2]{*}{$Acc_{judge}$} & Source Model & 91.0  & 82.5  & 92.9  & 99.2  & 97.4  & 92.1  \\
          & Init Target Model in SFUDA & 52.2  & 71.3  & 69.9  & 56.9  & 89.4  & 51.6  \\
    \bottomrule
    \end{tabular}%
  \label{table_membership}%
\end{table}%

\vspace{-2em}

\subsection{Domain Adaptation Results}
\label{sec_results}
Table. \ref{table_digit} and Table. \ref{table_office_caltech} report the multi-source UDA accuracy on Digit-Five and Office-Caltech datasets, respectively. Each column represents a target domain. Table. \ref{table_office_31} and Table. \ref{table_office_home} report the single-source UDA accuracy between each two domains on Office-31 and Office-Home datasets, respectively. It can be seen that SFUDA achieves higher accuracy than Source Only in average accuracy, whether for single-source or multi-source. The average accuracy of SFUDA is also close, and sometimes exceeds other methods that taking advantage of model transferring under source-free UDA setting or even UDA setting. This confirms that queries based on another dataset and adversarial examples generated by DAT have indeed transferred supervision information from source model.

Table. \ref{table_domainnet} compares multi-source UDA accuracy between SFUDA and the other methods on DomainNet dataset. Due to the large gap between domains and a total of 345 categories, DomainNet is the most challenging of all the datasets in this paper. This is verified by the significant performance degradation on the other strategies CP and GNP. SFUDA achieves an average accuracy of 35.7\% on the DomainNet, which has surpassed the FADA method and is close to the Source Only performance in the `real' target domain. Considering that the SFUDA method only uses a another dataset totally irrelevant to the source and target data to approximate the target model via ing, and no source or target data is used during the entire  process, this result is promising.

\subsection{Ablation Study and Membership Inference Attack}

In Table. \ref{table_ablation_study}, we compare the influence of three supervision information to SFUDA's performance. $L_{thrid-party}$ represents the performance of target model initialized by another dataset, $L_{DAT}$ represents the performance of retrained target model using adversarial examples generated by DAT, $L_{target}$ denotes the performance of target model after fine-tuning on target data under self-supervised pseudo-labeling. From the average accuracies of multi-source domain adaptation on different datasets, all three supervision information have improved the performance of SFUDA.

We further verify the impact of model transferring on source data privacy with membership inference attack. According to the setting in \cite{shokri2017membership}, we select a source model $M_{S_a}$ from source domain $\mathbb{D}_{S_a}$ as the shadow model to generate attack dataset $D_{atk} = \{x_{atk}, y_{atk}\}_{j=1}^{n_{atk}}$. $x_{atk}$ is a vector of dimension $K$, which represents the soft label of data in $K$ categories after passing through $M_{S_a}$. We input the entire dataset into the shadow model to get the soft label, i.e., the logits after softmax layer. $y_{atk}$ is a binary label, indicating whether data exist in the training set of $M_{S_a}$. We label the confidence on shadow model's training data as 1, indicating that the model has been trained on these data. Confidence on other data are labeled as 0. We use the attack dataset to train an attack model $M_{atk}$ based on 5-layer FCN, so that the attack model obtain the ability to judge whether the data is in the training set of a model according to the confidence distribution on the data.

After the attack model $M_{atk}$ is trained, we test it on source model $M_{S_b}$ from another source domain $\mathbb{D}_{S_b}$. Similarly, we input the entire dataset into the source model and calculate the accuracy of $M_{atk}$'s judgement. We also compare the judgement accuracy of the $M_{atk}$ on the target model of our SFUDA method initialized by another dataset. In Table. \ref{table_membership}, we report the judgement accuracy $Acc_{judge}$ of $M_{atk}$ on each source model of DomainNet and corresponding SFUDA model initialized with ImageNet. It can be seen that $Acc_{judge}$ of attack model has a significant drop in the initialized model of SFUDA compared to the source model. In other words, the method of obtaining source domain information through a another dataset of SFUDA can effectively resist the potential threat of membership inference attack in data-critical domain adaptation scenarios.

\section{Conclusion}
In this paper, we explore and source-free UDA setting that source and target domain models cannot be transferred. We propose SFUDA method to obtain supervision information from the source model taking advantage of another dataset. We further propose distributionally adversarial training to align distribution between another data with target data for more informative  results. Experimental results on 5 domain adaptation datasets demonstrate that SFUDA achieves comparable accuracy without transferring of source or target models. We further verify the advantage on data security of SFUDA with membership inference attack.


\bibliographystyle{IEEEtran}
\bibliography{neurips_2021}

\end{document}